\documentclass[10pt,twocolumn,letterpaper]{article}

\usepackage{cvpr}
\usepackage{times}
\usepackage{epsfig}
\usepackage{graphicx}
\usepackage{amsmath}
\usepackage{amssymb}

\usepackage[pagebackref=true,breaklinks=true,letterpaper=true,colorlinks,bookmarks=false]{hyperref}

 \cvprfinalcopy 

\def\cvprPaperID{****} 

\ifcvprfinal\fi
\begin{document}

\title{Multitask Text-to-Visual Embedding with Titles and Clickthrough Data}

\author{Pranav Aggarwal, Zhe Lin, Baldo Faieta, Saeid Motiian\\
Adobe Inc.\\
San Jose, USA\\
{\tt\small \{pranagga, zlin, bfaieta, motiian\}@adobe.com}
}

\maketitle

\begin{abstract}
Text-visual (or called semantic-visual) embedding is a central problem in vision-language research. It typically involves mapping of an image and a text description to a common feature space through a CNN image encoder and a RNN language encoder. In this paper, we propose a new method for learning text-visual embedding using both image titles and click-through data from an image search engine. We also propose a new triplet loss function by modeling positive awareness of the embedding, and introduce a novel mini-batch-based hard negative sampling approach for better data efficiency in the learning process. Experimental results show that our proposed method outperforms existing methods, and is also effective for real-world text-to-visual retrieval.  
\end{abstract}

\section{Introduction}

Image search has been a well-studied problem in both research and industry. With the recent advancement in deep learning, cross-modal retrieval between texts and images has been a central problem in the field of language and vision. Previous methods mostly rely on tags extracted from textual data or automatically inferred from images to perform text-based image retrieval. However, these methods are prone to errors when the text query is long due to the lack of flexibility on variations of language descriptions.

Text-visual embedding (also called semantic-visual embedding) aims to map text and visual information to the same feature space so that cross-modal retrieval can be performed by nearest neighbor search in the feature space. It can effectively cope with the limitations of bag-of-words based models common to many image search algorithms. Text-visual embedding is particularly effective for long text queries with proper encoding of language information. In recent developments, generative models are used to get image representations for text and then perform nearest neighbours while some methods try to perform operations in image feature extraction side to achieve the common vector space. Instead we try to build a shallow architecture, which tries to get only the text in a space that can be extracted very easily, such as the ResNet \cite{resnet} or VGG \cite{Simonyan14c} feature space. 

In this paper, our contributions include: (1) We propose a new way of getting images and text into a common space by performing multi-task training using a combination of click-through dataset for user intention and image caption dataset to equip our model for long sentences and remove user query noise, while keeping the image feature extraction architecture unchanged. Using this combinations helps our model tackle real-world text queries. (2) We also propose a novel loss function that tries to combine the advantages of the contrastive loss and triplet loss, we call it positive aware triplet ranking loss. (3) This paper describes how we can select ``hard'' negatives which will eventually decide the kind of generalization we see in the image or text results by influencing the amount of tightening of the entity clusters.

\begin{figure}[t]
\begin{center}
   \includegraphics[width=0.8\linewidth]{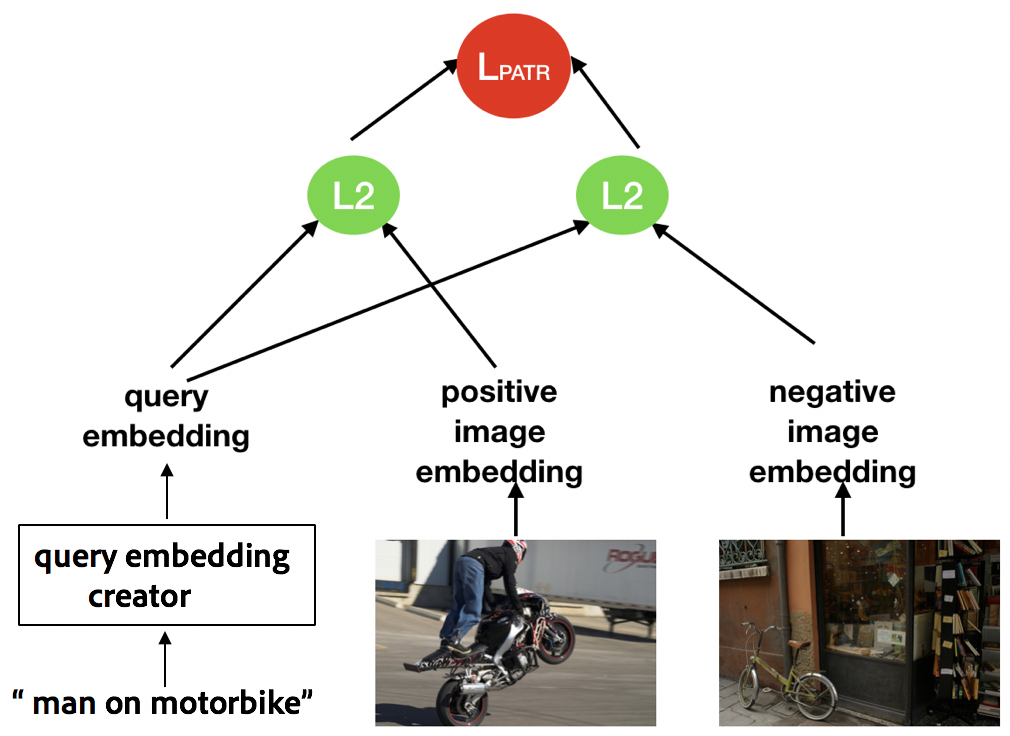}
\end{center}
   \caption{Conceptual illustration of our training mechanism using one sample: We perform this for title dataset batch and click-through dataset
batch separately and the average of the two losses is then back-propagated
through the query embedding creator.}
\label{fig:long}
\label{fig:onecol}
\end{figure}
\section{Related work}
The problem described above is part of a broader concept called Metric learning which has been studied in several fields such as machine learning~\cite{Distance2009Weinberger}, information retrieval~\cite{mcfee2010metric}, and computer
vision~\cite{Motiian_2017_ICCV}. The goal of a metric
learning algorithm is to project samples from two different domains/modals to a common latent space
so that similar data samples
(e.g. from the same class and different domains/modals) lie close to each other and dissimilar
data samples (e.g. samples from different classes) lie far away
from each other.

In this paper we consider image and text as two modals of the data. Canonical Correlation
Analysis (CCA)~\cite{hardoon2004} is one of the early methods to find a common latent space between text and image. \cite{ranjan2015multi} extends CCA for learning a common space taking into account the high level semantic information in the form of multi-label annotations. Similar to most traditional machine learning algorithms, CCA has some intrinsic limitations where it may not be suitable for large scale datasets. \cite{yan2015deep} combines deep learning with CCA and addresses the problem of matching images
and captions in a joint latent space learnt with deep canonical correlation analysis (DCCA). For metric learning, a deep network can learn a complex non-linear projection function for each modal of the data using coupled networks~\cite{karpathy2014deep,liong2017deep,mithun2018webly}.

Recently, \cite{nawaz2018revisiting} goes beyond the coupled networks and designed a cross modal retrieval approach with only one network. This is done by fusing images and texts before passing to a single network.  \cite{gu2018look} significantly improves the state-of-the-art performance
on cross-modal image-caption retrieval by using the coupled structure together with two additional generative networks. Adversarial learning also used in~\cite{sahcross} to find a common space using within-modality and across modality discriminators.

\section{Our Text-Visual multi-task training network}

\subsection{Creating the common embedding space}
First we need to fix the embedding space for which we would like our images and text to be projected to. For this we make use of already existing pre-trained architectures that are trained on a large corpus of images to predict tags such as VGG-19 \cite{Simonyan14c} , ResNet-152 or ResNet-50 \cite{resnet} models. We select the layer before the last activation layer (Softmax) as the common embedding space.

\subsection{Text embedding creation network}
\begin{figure}[t]
\begin{center}
   \includegraphics[width=1.0\linewidth]{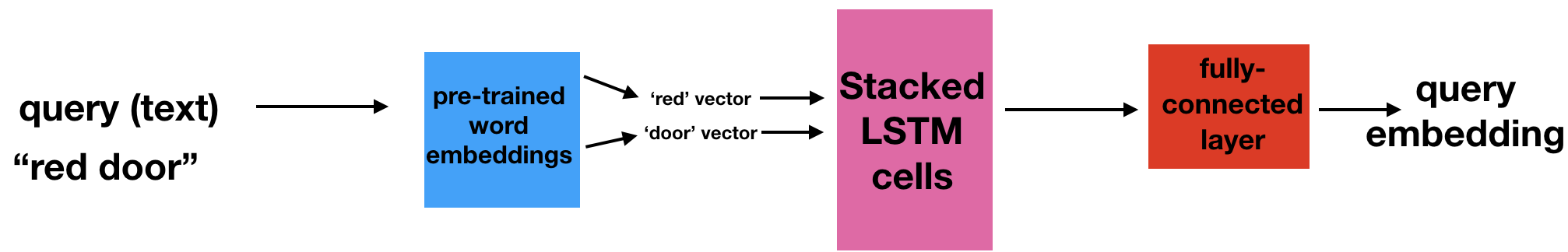}
\end{center}
   \caption{Query embedding creator architecture: This architecture represents the query embedding creator block seen in Figure 1.}
\label{fig:long}
\label{fig:onecol}
\end{figure}

Given a text input to the network, we first associate the vector representation for each word. We make use of pre-trained embeddings such as Fasttext embeddings \cite{grave2018learning} to get the vector representations of the words which produces a 300 sized vector for each word. The sequential information is then captured using the Long Short Term Memory units \cite{lstm} (LSTM). Stacking the LSTMs and adding a dropout \cite{dropout} to the LSTM cells helps in improving the performance of the model. Output of the last LSTM unit is then given to a fully connected layer which tries to match the dimension space of the LSTM output to image embeddings.

\subsection{Hard negative image selection}

We perform a multi-task training in which we train separately from both the title dataset and click-through dataset. For both cases, we have the input text associated with a positive image. For the title dataset, the text is the title and the positive image is the image associated with the title. For the click-through dataset the text is the text of the query and the positive image is the image being clicked. As for the negative images, we select images within the batch. We have to make sure that the negative images are similar enough to the positive image so as to increase discrimination by the network but selected in such a way that they don't belong to the same category. Steps to sample negative images:

{\bf 1:} We use a batch of 512 image embedding samples, where each sample corresponds to either the clicked image associated to a text query for the click-through dataset or the image associated with a title for the title dataset. These are the positive image embeddings. 

{\bf 2:} For every positive sample in the batch, 

{\bf 2.1:} We calculate the square distance to every other sample in the batch

{\bf 2.2:} We remove those samples for which its associated text shares any words in common with those of the positive sample, excluding any stop words. So, for example, if the text query associated with the positive sample is "man on a motorbike", the words we consider are "man" and "motorbike" and we remove those samples that have either "man" or "motorbike" in their associated text. If we want to select harder negatives, we only remove those samples that have both "man" and "motorbike" instead.

{\bf 2.3:} From the samples left, we select N negative samples with the least squared distance from the positive sample.

\subsection{Positive aware triplet ranking loss}
Once we find the squared distance between the positive image embedding and the query embedding ($s_p$) and the squared distance between the negative image embedding and the query embedding ($s_n$), we find the positive aware triplet ranking loss:	
    \begin{equation}
             L_{PATR} = s_p + max(0, \eta - s_n)
    \end{equation}

here $\eta$ tries to penalize $s_n$, therefore higher the $\eta$, tighter are the clusters. In triplet loss \cite{triplet2010}\cite{triplet2011}\cite{triplet2014} given below:
    \begin{equation}
			L_{triplet} = max(s_p - s_n + \rho, 0)  
    \end{equation}
here $\rho$ acts like the $\eta$. When we consider $s_p$ inside the max(), the loss function tries to minimize the $s_p$ $-$ $s_n$ by increasing or decreasing the value of both $s_p$ and $s_n$ together, with $s_p$ getting impacted more, causing the difference to automatically increase. We wanted to consider both these values separately in our loss function. $L_{PATR}$ tries to minimize $s_p$ and also separately tries to maximize $s_n$. This causes the positive image embeddings to become very similar to the text query embeddings, therefore forcing them to lie in the same cluster while tightening these clusters by maximizing $s_n$. The effectiveness can be increased by adding multiple negative samples by selecting the top N negative samples. Then our loss function becomes: 
    \begin{equation}
        L_{PATR'} = s_p + \sum_{i=1}^{N} (max(0, \eta - s_{n_i}))  
    \end{equation}
Our final multi-task training loss is:
\begin{equation}
        L_{our} = (L_{{PATR'}_{caption}} + L_{{PATR'}_{click}})/2
\end{equation}
\section{Experimental results}
\subsection{Datasets}
Pascal Sentence Dataset \cite{PascalSentence} : The dataset contains 1K images with 5 captions each. To be fair we use the same dataset partition as seen in \cite{CCL2017} i.e. 800 images for training and 200 for testing. The images are already divided into 20 categories. We use the category information only during testing with minor obvious mislabelling corrections.

Adobe Stock dataset: We collect 1M user typed text query-clicked image pairs using our current Adobe Stock search engine and 1M caption (title)-image pairs provided by Adobe Stock users, for training. The testing dataset consists of 5K caption-image pairs and 5k user typed text query-clicked image pairs for which we find the evaluations separately and then show the average. The user provided captions are not always very elaborate and precise. We find results with this dataset to show that our loss function coupled with multi-task training out-performs other loss functions and methods in a real-world case scenario.

\subsection{Implementation}
For all experiments we use Tensorflow \cite{Tensorflow2016} and Adam Optimizer \cite{kingmab14} with decay parameter (beta1=0.99) to train our model. Also we stack the LSTMs 5 times with 25\% of the nodes dropped-out. To get image features from VGG architecture we extract the relu7 features (4096 dim) and from ResNet architectures we extract the pool5 layer (2048 dim).

\begin{table}
\centering
\caption{Training hyper-paramters and image features}
\begin{tabular}{ |l|l|l|l|l| } 
 \hline
 Dataset & No.of  & learning & LSTM  & Image\\
   & epoch & rate & units & feature \\ \cline{2-2}
 \hline
 Pascal & 50 & 0.0005, & 18 & VGG-19, \\
 Sentence & &0.0001 & & ResNet-152 \\ \cline{2-2}
 \hline
 Adobe  & 30  & 0.0001 & 15 & ResNet-50\\
 Stock & & & & \\
\hline
\end{tabular}

\end{table}

For the Pascal Sentence Dataset we implement our loss function (3) with a margin of 1.0 and as we only have caption data for this experiment so we do not implement the multi-task training. The number of top hard negative samples (N) taken is 3.

For Adobe Stock Dataset we compare our loss function (4) ($\eta$ = 1.2) with multi-task triplet loss (2) ($\rho$ = 0.5) and l2 loss (Baseline). To show fair comparisons we use only one negative sample for both $L_{triplet}$ and $L_{PATR'}$. We also compare our multi-task implementation (4) with only click-trough dataset and caption dataset implementation using equation (1). 

\subsection{Evaluation Metric}
We report the mean Average Precision (mAP) score described in \cite{mAP2014} for Pascal Sentence dataset. Considering we have first R top-ranked retrieved data samples for each query, we can define AP for each sample as:

\begin{equation}
             mAP = \frac{1}{M} \sum_{r=1}^{R} p(r).rel(r)
    \end{equation}
here M is the number of relevant samples retrieved. p(r) is the precision at r and rel(r) is 1 if r is relevant and 0 if not. The retrieved data is considered as relevant if it has the same semantic label as the query. mAP is the average of all the samples. We report mAP@50 (R=50).

As we do not have semantic labels for the Adobe Stock dataset so we use the measure in \cite{orderemb2016} for our evaluation i.e., R@K which is defined as the percentage of text queries in which the ground-truth images are contained in the first
K retrieved results.

\subsection{Cross Modal Retrieval}
\begin{figure}[t]
\begin{center}
   \includegraphics[width=0.9\linewidth]{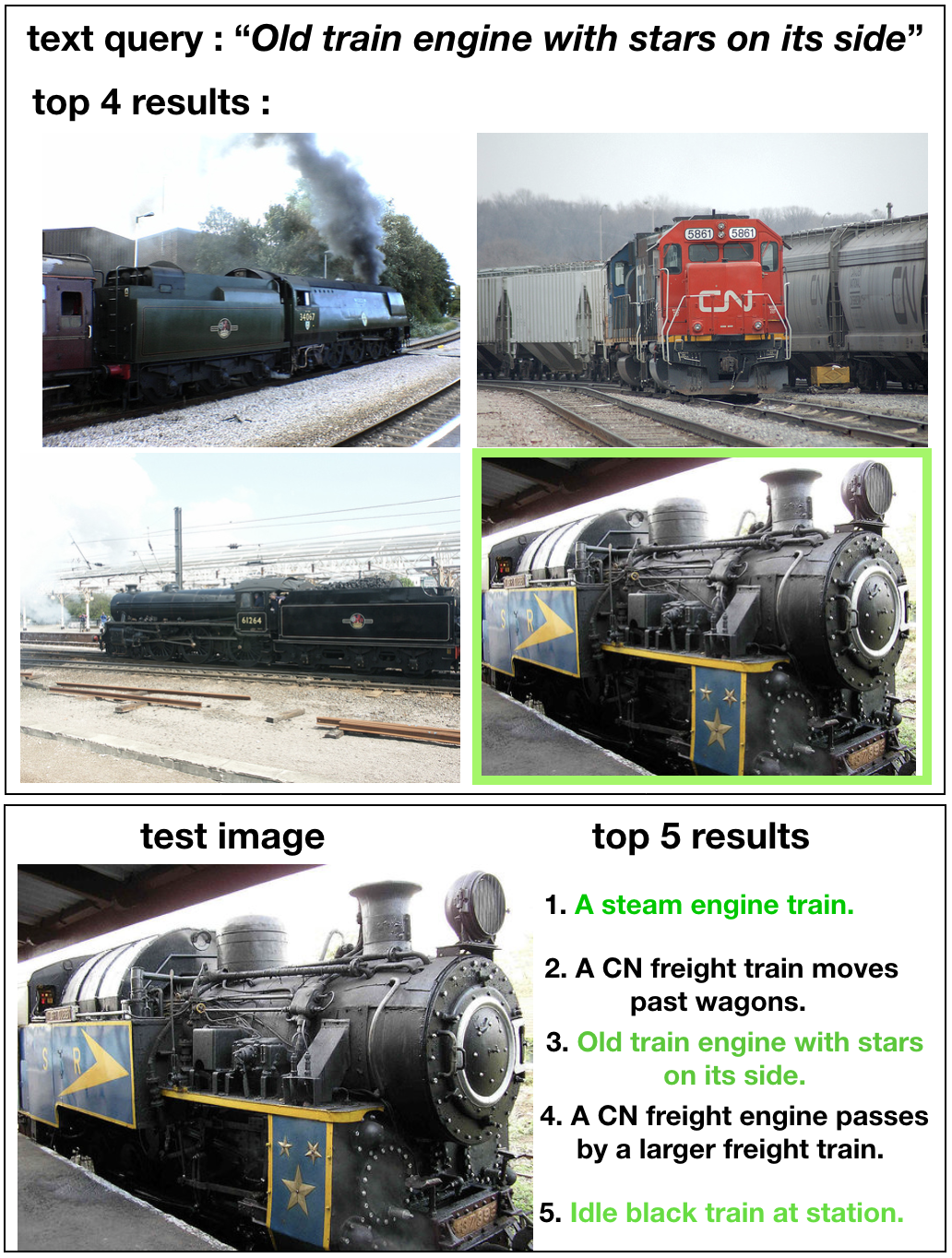}
\end{center}
   \caption{Cross-modal retrieval results using Pascal Sentence Dataset and our method (ResNet-152) (green color represents ground truth). Top box shows Txt2Img results while bottom box shows Img2Txt results.}
\label{fig:long}
\label{fig:onecol}
\end{figure}

\begin{table}
\centering
\caption{mAP evaluation using Pascal dataset}
\begin{tabular}{ |l|c|c| } 
 \hline
 Method & Img2Txt & Txt2Img \\
 \hline
 Corr-AE \cite{mAP2014} & 0.290 & 0.279 \\
 \hline
 ACMR \cite{adversarial2014} & 0.535  & 0.543 \\
\hline
 CVS \cite{CVS2018} & 0.604  & 0.578 \\
\hline
Ours (VGG-19)   & 0.634 & 0.533 \\
\hline
Ours (ResNet-152)  & \textbf{0.656} & 0.562 \\
\hline
\end{tabular}

\end{table}

In Table 2 we see that our methods easily out-perform others in Img2Txt retrieval. The Txt2Img evaluation using ResNet-152 features shows almost state of the art performance but can be improved by having larger dataset as we have more semantically similar images to select the negative samples from.

\begin{table*}[t]
\caption{Ranking evaluations for different loss functions using both caption and click-through Adobe Stock dataset}
\centering
\begin{tabular}{ | l| c| c| c| c| c| c| c| c| c|} 
 \hline
  & \multicolumn{2}{c}{Caption Dataset} & & \multicolumn{2}{c}{Click-through Dataset} & & \multicolumn{2}{c}{Average} & \\
 \hline
  Method & R@1 & R@10 & R@20 & R@1 & R@10 & R@20 & R@1 & R@10 & R@20 \\
 \hline
 l2  & 0.118 & 0.430 & 0.546 & 0.071 & 0.286 & 0.371 & 0.094 & 0.358 & 0.459 \\
\hline
 triplet loss \cite{triplet2010}  & 0.153	& 0.485 & 0.583 & 0.087 & 0.315 & 0.398 & 0.120 & 0.400 & 0.491 \\
\hline
 Ours   & \textbf{0.174} & \textbf{0.511} & \textbf{0.620} & \textbf{0.092} & 0.312 & 0.393 & \textbf{0.133} & \textbf{0.412} & \textbf{0.506} \\
\hline
\end{tabular}
\end{table*}

\begin{table*}[t]
\caption{Ranking evaluations for different training combinations of caption and click-through dataset with our loss function}
\centering
\begin{tabular}{ | l| c| c| c| c| c| c| c| c| c|} 
 \hline
  & \multicolumn{2}{c}{Caption Dataset} & & \multicolumn{2}{c}{Click-through Dataset} & & \multicolumn{2}{c}{Average} & \\
 \hline
  Training Dataset & R@1 & R@10 & R@20 & R@1 & R@10 & R@20 & R@1 & R@10 & R@20 \\
 \hline
 only clicks & 0.076 & 0.322 & 0.425 & 0.091 & 0.312 & 0.404 & 0.084 & 0.317 & 0.414 \\
\hline
 only titles & 0.180 & 0.514 & 0.619 & 0.052 & 0.193 & 0.262 & 0.116 & 0.354 & 0.441 \\
\hline
 clicks and titles & 0.174 & 0.511 & \textbf{0.620} & \textbf{0.092} & 0.312 & 0.393 & \textbf{0.133} & \textbf{0.412} & \textbf{0.506} \\
\hline
\end{tabular}
\end{table*}

\subsection{Real-World Analysis for Image Retrieval}
In Table 3 we see that our method is able to retrieve the ground-truth images with better performance for all the R cases for kinds of Adobe Stock dataset as compared to the other loss functions. 

Table 4 shows that using multi-task training with both click-through and caption (title) dataset can combine the advantages of the methods when trained only with one type of dataset. Therefore we see a high increase in the average values for all the R cases when using our approach.

\begin{figure}[t]
\begin{center}
   \includegraphics[width=1\linewidth]{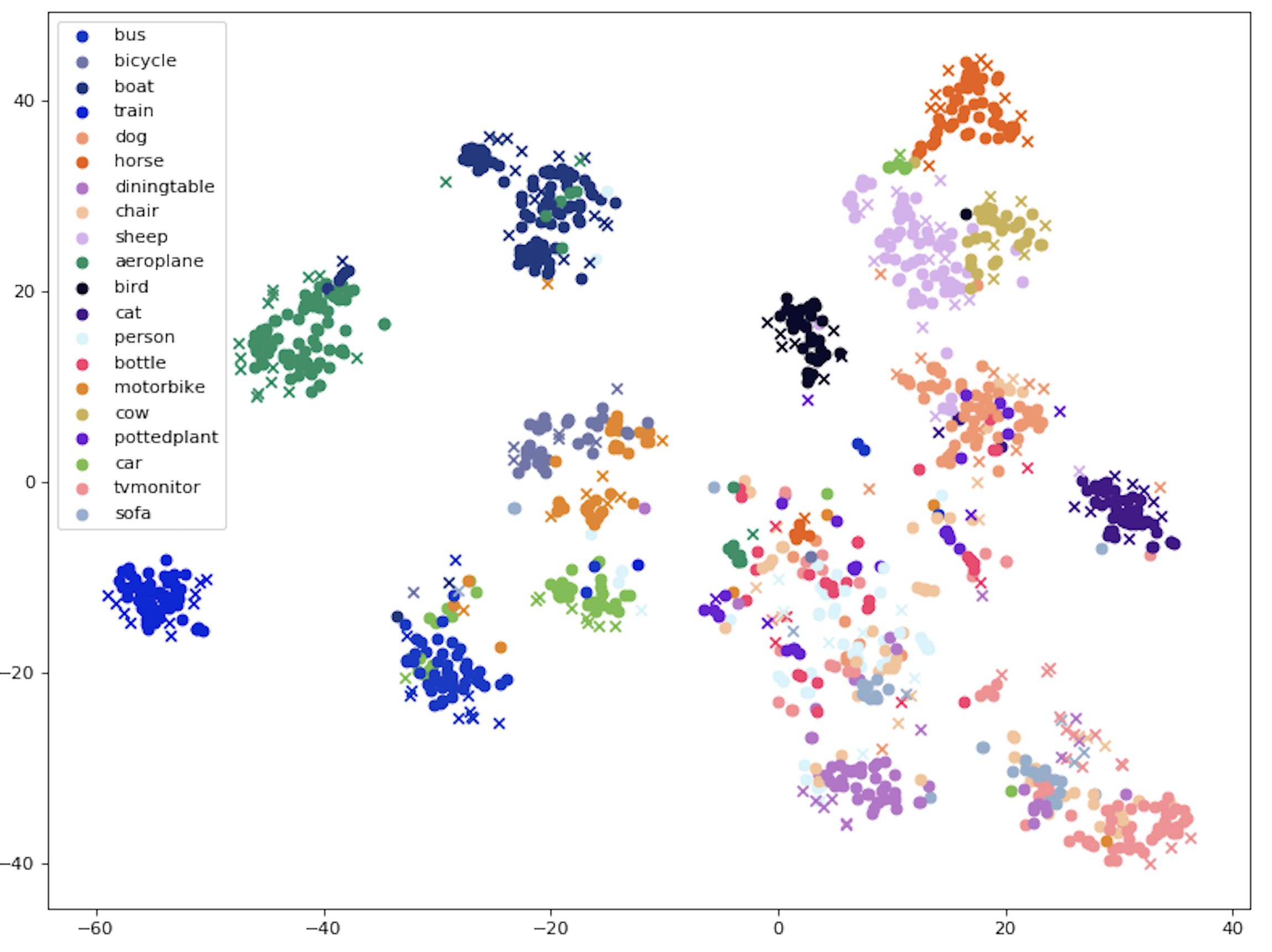}
\end{center}
   \caption{Category level T-SNE clustering of Pascal Sentences testing dataset using our method (ResNet-152). Here ``x'' represents image and ``o'' represent text. (Please zoom in for better visualization).}
\label{fig:long}
\label{fig:onecol}
\end{figure}

\section{Conclusions}

We present a shallow deep learning architecture along with a novel loss function which tries to get the text into the same vector space as that of image features. This model takes into account the practical use cases of a search engine and therefore we show how multi-task training involving user click data and caption data can help get better results for real world use cases. Also we demonstrate how negative samples can be an important aspect of how we get our query clusters. In Figure 4. we see that the text captions get mapped into the visual space. One good example seen here is that the bicycle text (sentence) cluster is very close to the motorbike text cluster (without much overlap) as their objects are visually similar therefore gaining visual intuition.

{\small
\bibliographystyle{ieee_fullname}
\bibliography{cvpr_final}
}
\end{document}